\documentclass[]{article}

\usepackage[final]{nips_2018}
\usepackage[utf8]{inputenc} 
\usepackage[T1]{fontenc}    
\usepackage{hyperref}       
\usepackage{url}            
\usepackage{booktabs}       
\usepackage{amsfonts}       
\usepackage{nicefrac}       
\usepackage{microtype}      
\usepackage{amsmath}
\usepackage[pdftex]{graphicx}

\DeclareMathOperator*{\argmin}{arg\,min}

\title{Multi-view Factorization AutoEncoder with Network Constraints for Multi-omic Integrative Analysis}

\author{
	Tianle Ma \\
	Department of Computer Science\\
	University at Buffalo\\
	Buffalo, NY 14260 \\
	\texttt{tianlema@buffalo.edu} \\
	\And
	Aidong Zhang\\
	Department of Computer Science\\
	University at Buffalo\\
	Buffalo, NY 14260 \\
	\texttt{azhang@buffalo.edu} \\}

\begin{document}

\maketitle

\begin{abstract}
Multi-omic data provides multiple views of the same patients.
Integrative analysis of multi-omic data is crucial to elucidate the molecular underpinning of disease etiology. However, multi-omic data has the ``big p, small N'' problem (the number of features is large, but the number of samples is small), it is challenging to train a complicated machine learning model from the multi-omic data alone and make it generalize well. Here we propose a framework termed Multi-view Factorization AutoEncoder with network constraints to integrate multi-omic data with domain knowledge (biological interactions networks). Our framework employs deep representation learning to learn feature embeddings and patient embeddings simultaneously, enabling us to integrate feature interaction network and patient view similarity network constraints into the training objective. 
The whole framework is end-to-end differentiable. We applied our approach to the TCGA Pan-cancer dataset and achieved satisfactory results to predict disease progression-free interval (PFI) and patient overall survival (OS) events. Code will be made publicly available.
\end{abstract}

\section{Introduction}
Many applications have multi-view data. Notably, massive amounts of patient data with multi-omic profiling have been accumulated during the past few years. For example, TCGA network \cite{hutter2018cancer} have generated comprehensive multi-omic molecular profiling for more than 10,000 patients from 33 cancer types. Each type of -omic data (e.g., genomic, transcriptomic, proteomic, and epigenomic, etc.) represents one view from the same set of patients. Each view has a different feature set (for example, gene features, miRNA features, protein features etc.) and can provide complementary information for other views. Integrative analysis of multi-omic data is important for predicting cancer (sub)types and disease progression, but is very challenging. Currently most results generated by TCGA network are mainly based on statistical analysis, though machine learning approaches are increasingly popular to tackle the problems.

Meanwhile, in the past decade, deep learning brought about significant breakthroughs in computer vision, speech recognition, natural language processing and other fields \cite{lecun2015deep}. However, conventional deep learning models requires massive training data with clearly defined structures (such as images, audio, and natural languages), and are not suitable for multi-omic integrative analysis.

In this work, we propose a model termed Multi-view Factorization AutoEncoder (MAE), which combines the ideas from multi-view learning \cite{zhao2017multi} and matrix factorization \cite{Bell2009} with deep learning, in order to utilize the great representation power in deep learning models. The backbone of a Multi-view Factorization AutoEncoder model consists of multiple autoencoders (one for each view) as submodules, and a submodule to combine multiple views for supervised learning.

To alleviate overfitting and overcome the problem of ``big $p$, small $N$'' problem, we incorporate molecular interaction networks as graph constraints into our training objective. These feature interaction networks are derived from public knowledgebases, thus enabling our model to incorporate domain knowledge.

Since all views are from the same set of patients, the patient similarity networks derived from these learned views should share some information, too. As a result, in addition to feature interaction network constraints, we also added patient similarity network constraints to our training objective. Our model equipped with feature interaction network and patient similarity network constraints performs better than traditional machine learning methods such as SVM and Random Forest on the TCGA Pan-cancer dataset \cite{hutter2018cancer}.

\section{Related work}
Multi-omic data analysis has been a hot topic in cancer genomics \cite{hutter2018cancer,ebrahim2016multi,henry2014omictools}. Most the work had been focused on comprehensive molecular characterization of individual cancer types \cite{hutter2018cancer,shen2018integrated}, which mainly employed statistical analysis of molecular features associated with clinical outcomes. Machine learning approaches also have been applied to study individual -omic data types \cite{malta2018machine} and integrate multi-omic data \cite{way2018machine,angione2016multiplex}. These approaches mainly employ traditional machine learning techniques, for example, logistic regression \cite{malta2018machine}, random forest \cite{way2018machine}, and similarity network fusion \cite{angione2016multiplex}. 

Many machine learning approaches for multi-omic data analysis fall into the category of unsupervised clustering, such as iCluster \cite{shen2012integrative}, SNF \cite{wang2014similarity}, ANF \cite{ma2017integrate}, etc. These approaches are either based on probabilistic models \cite{shen2012integrative} or network-based regularization \cite{hofree2013network}. 
While deep learning approaches had been applied to sequencing data \cite{alipanahi2015predicting} \cite{bovza2017deepnano}, imaging data \cite{wang2016deep}, medical records \cite{pham2016deepcare}, and other individual data types, few focused on integrating multi-omic data. 
Co-training, co-regularization and margin  consistency approaches have been developed for multi-view learning \cite{zhao2017multi}, while integrating deep learning with multi-view learning is still an open frontier \cite{zhao2017multi}. 

Our work is closely related to multi-modality deep learning \cite{baltruvsaitis2018multimodal}, which had been successfully applied to combine audio and video features\cite{ngiam2011multimodal} by employing shared feature representations. In addition to integrating multi-modality data, our proposed approach can learn feature representations and object (patient) embeddings simultaneously, which enables us to integrate feature interaction networks as domain knowledge, as well as to enforce view similarity network constraints in the training objective.
 
Besides adding regularizers into training objectives, another way to incorporate biological networks into the model is to directly encode biological networks into the model architecture \cite{hu2018deep,ma2018using}. However, these approaches usually require using subcellular hierarchical molecular networks, which we do not have high-quality data available for humans (though a few datasets are available for simple organisms such as bacteria). Given the fact that most human biological interaction networks such as protein-protein interaction networks are highly incomplete and noisy, adding network regularizers to the training objective instead of directly encoding the noisy interaction network into the model architecture provides more flexibility and alleviates the risk of adopting potentially wrong model architecture.

Our work is also related matrix factorization \cite{Bell2009} and autoencoder, which will be reviewed briefly in the next section along with the detailed description of our proposed method.

\section{Multi-view Factorization Encoder}

\paragraph{Notations} Suppose there are $N$ samples, $V$ types of -omic data. We refer to each -omic data type as a view.

We represent $V$ types -omic data using a series of sample-feature matrices: $\mathbf{M}^{(i)} \in \mathbb{R}^{N \times p^{(i)}}, i=1,2,\cdots, V$. 
$p^{(i)}$ is the feature dimension for view $i$.

In the following, we first describe a framework for a single view, then describe how to integrate multiple views. When describing a single view, we drop the superscript $^{(\cdot)}$ for simplicity. When describing a matrix $\mathbf{M}$, we use $M_{ij}$ to represent the element of $i$th row and $j$th column, $\mathbf{M}_{i,\cdot}$ to represent the $i$th row vector, and $\mathbf{M}_{\cdot, j}$ to represent $j$th column vector.

Let $\mathbf{M} \in \mathbb{R}^{N \times p}$ be a sample-feature matrix, with rows corresponding to $N$ samples and columns $p$ features. These $p$ features are not independent. We can represent the interactions among these $p$ features with a graph $\mathbf{G} \in \mathbb{R}^{p \times p}$. For example, if the $p$ features are protein expressions, then $\mathbf{G}$ can be a protein-protein interaction network, which can be obtained from public knowledgebases, such as STRING \cite{szklarczyk2014string}, Reactome \cite{croft2013reactome}, etc. $\mathbf{G}$ can be a weighted graph with non-negative elements, or an unweighted graph with elements being either 0 or 1.

Denote $\mathbf{L}_{G}=\mathbf{D} - \mathbf{G}$ as the graph Laplacian of $\mathbf{G}$ ($\mathbf{D}$ is a diagonal matrix with $D_{ii}=\sum_{j=1}^{p} G_{ij}$).

\subsection{Low-rank matrix factorization}
Matrix factorization \cite{Bell2009} and its variants are commonly used for dimensional reduction and clustering. It is often a reasonable assumption that $\mathbf{M}$ has low rank in many real world applications. In order to identify the underlying sample clusters, low-rank matrix factorization can be applied to $\mathbf{M}$:

\begin{center}
	$\mathbf{M} \approx \mathbf{X} \mathbf{Y}$,
	where $\mathbf{X} \in \mathbb{R}^{N \times k}, \mathbf{Y} \in \mathbb{R}^{k \times p}, k<p$
\end{center}

In order to find a good solution $\{\mathbf{X}, \mathbf{Y}\}$, some constraints are usually added as regularizers in the objective function or enforced in the learning algorithm. For example, if $\mathbf{M}$ is non-negative matrix (e.g., gene feature count matrix), Non-negative Matrix Factorization (NMF) \cite{lee2001algorithms} is often used to ensure both $\mathbf{X}$ and $\mathbf{Y}$ are non-negative.

In general, the objective functions can be formulated as follows:

\begin{equation}\label{eq:mf_objective}
	\argmin_{\mathbf{X}, \mathbf{Y}} \left\Vert \mathbf{M} - \mathbf{X} \mathbf{Y} \right\Vert^2_F + \lambda R(\mathbf{X}, \mathbf{Y})
\end{equation}

$R(\mathbf{X}, \mathbf{Y})$ is a regularizer for $\mathbf{X}$ and  $\mathbf{Y}$. For example, $R(\mathbf{X}, \mathbf{Y})$ can include $L_2$ and $L_1$ norms for $\mathbf{X}$ and $\mathbf{Y}$. More importantly, structural constraints based on biological interaction networks can also be incorporated into $R(\mathbf{X}, \mathbf{Y})$, which will be discussed later.

\paragraph{Interpretation} Suppose there are $k$ factors that fully characterize these samples. $\mathbf{X} \in \mathbb{R}^{N \times k}$ can be seen as a sample-factor matrix. These $k$ factors are not directly observable. Instead, we observed $\mathbf{M} \in \mathbb{R}^{N \times p}$, which can be seen as a linear transformation of $\mathbf{X}$.
And $\mathbf{Y} \in \mathbb{R}^{k \times p}$ can be seen as the matrix of such a linear transformation from $\mathbb{R}^{k}$ to $\mathbb{R}^{p}$. The $k$ rows of $\mathbf{Y}$ can be seen as a basis for the underlying factor space. Therefore $\mathbf{M}$ is generated by a linear transformation $\mathbf{Y}$ from $\mathbf{X}$, the inherent non-redundant representation of $N$ samples. In a sense, this formulation can be seen as a shallow linear generative model.

\paragraph{Limitations} The limitations of this simple matrix factorization model arise from its shallow linear structure. The representation power of linear models is very limited. In most cases, the transformations are non-linear. To increase the model representation capacity, we will discuss nonlinear factorization with multi-layer neural networks, which can approximate any complex nonlinear transformations with sufficient data. 

\subsection{Non-linear factorization with AutoEncoder}
Instead of direct matrix factorization, we can use an autoencoder to reconstruct the observable sample-feature matrix $\mathbf{M}$.
While direct matrix factorization -- which can be seen as a one-layer autoencoder -- is limited to model nonlinear relationships, multi-layer autoencoder can approximate complex nonlinear transformations well. 

We use a multi-layer neural network with parameter $\mathbf{\Theta_{e}}$ as the encoder:

\begin{equation} \label{eq:encoder}
	Encoder(\mathbf{M}, \mathbf{\Theta_{e}}) = \mathbf{X} \in \mathbb{R}^{N \times k}
\end{equation}

Again $\mathbf{X}$ can be seen as a non-redundant factor matrix that contains essential information for all $N$ samples. We are using a multi-layer neural network to transform the observable sample-feature matrix $\mathbf{M}$ to its latent representation $\mathbf{X}$. 

The decoder is a transformation from latent factor space to the reconstructed feature space.
 
\begin{equation} \label{eq:decoder}
Decoder(\mathbf{X}, \mathbf{\Theta_{d}}) = \mathbf{Z} \in \mathbb{R}^{N \times p}
\end{equation}

As the entire autoencoder is a multi-layer neural network, we can arbitrarily split it into the encoder and the decoder components. For the convenience of incorporating biological interaction networks into the framework, we make the encoder (Eq.~\ref{eq:encoder}) contain all layers but the last one, and the decoder only contain the last linear layer. The parameter of decoder (Eq.~\ref{eq:decoder}) is simply a linear transformation matrix as in matrix factorization:

\begin{equation} \label{eq:decoder_matrix}
	\mathbf{\Theta_{d}} = \mathbf{Y} \in \mathbb{R}^{k \times p}
\end{equation}

Therefore the reconstructed signal is 
\begin{equation} \label{eq:simplified_autoencoder}
	\mathbf{Z} = Encoder(\mathbf{M}, \mathbf{\Theta_{e}}) \cdot \mathbf{Y} = \mathbf{X} \mathbf{Y}
\end{equation}

The reconstruction error can be calculated with Frobenius norm: 
$\left\Vert \mathbf{M} - \mathbf{Z} \right\Vert^2_F$.

This formulation is different from matrix factorization in that the encoder is a multi-layer neural network that can learn complex nonlinear transformations through backpropagation.
In addition, the output of the encoder $\mathbf{X}$ can be seen as the learned representations for $N$ samples, and $\mathbf{Y}$ can be seen as learned feature representations (we can regard the columns of $\mathbf{Y}$ as learned vector representations in $\mathbb{R}^{k}$ for $p$ features). With learned patient and feature representations, we can calculate patient similarity networks and feature interaction networks, and add network regularizers to our training objective.

\subsection{Incorporate biological knowledge as network regularizers} \label{sec:net_reg}
Let $\mathbf{G} \in \mathbb{R}^{p \times p}$ be the interaction matrix among $p$ genomic features. $\mathbf{G}$ can be obtained from biological knowledgebases such as STRING \cite{szklarczyk2014string} and Reactome \cite{croft2013reactome}. 

With the factorization autoencoder model, we can learn a feature representation $\mathbf{Y}$. Ideally this representation should be ``consistent'' with the biological interaction network of these features. 
We use graph Laplacian regularizer to ``punish'' the inconsistency between the learned feature representation $\mathbf{Y}$ and the feature interaction network $\mathbf{G}$:

\begin{equation} \label{eq:trace_y_L}
Trace(\mathbf{Y} \mathbf{L}_G \mathbf{Y}^T) = \frac{1}{2} \sum_{i=1}^{p} \sum_{j=1}^{p} G_{ij} \lVert \mathbf{Y}_{\cdot, i} - \mathbf{Y}_{\cdot, j} \rVert ^2
\end{equation}

In Eq.~\ref{eq:trace_y_L}, $\mathbf{L}_G$ is the graph Laplacian matrix of $\mathbf{G}$. 
$G_{ij} \ge 0$ can be regarded as a ``similarity'' (interaction) measure between feature $i$ and feature $j$. Each feature $i$ is represented as a $k$-dimensional vector $\mathbf{Y}_{\cdot,i}$. The Euclidean distance between feature $i$ and $j$ in the learned feature space is simply $\lVert \mathbf{Y}_{\cdot, i} - \mathbf{Y}_{\cdot, j} \rVert$. 
$Trace(\mathbf{Y} \mathbf{L}_G \mathbf{Y}^T)$ can be served as a surrogate for the loss measuring the inconsistency between learned feature representation $\mathbf{Y}$ and existing interaction network $\mathbf{G}$. To see this, let's consider a case where $\mathbf{Y}$ is highly inconsistent with $\mathbf{G}$: suppose whenever $G_{ij}$ is large (i.e., feature $i$ and $j$ are similar based on existing knowledge), $\lVert \mathbf{Y}_{\cdot, i} - \mathbf{Y}_{\cdot, j} \rVert^2$ is also large (i.e., feature $i$ and feature $j$ are very different based on learned representations). Then the loss $Trace(\mathbf{Y} \mathbf{L}_G \mathbf{Y}^T)$ consists of the terms $G_{ij} \lVert \mathbf{Y}_{\cdot, i} - \mathbf{Y}_{\cdot, j} \rVert^2$, which accounts for the level of inconsistency between learned feature representation and biological knowledge, will be large, too.

The objective function for the aforementioned factorization AutoEncoder model incorporating biological interaction networks through the graph Laplacian regularizer is as follows:

\begin{equation} \label{eq:mf_objective2}
	\argmin_{\mathbf{\Theta_{e}}, \mathbf{Y}} \left\Vert \mathbf{M} - \mathbf{Z} \right\Vert^2_F + \alpha \ Trace(\mathbf{Y} \cdot \mathbf{L}_G \cdot \mathbf{Y}^T)
\end{equation}

$\alpha
\ge 0$ is a hyperparameter to balance the reconstruction loss and the network regularization term. To ensure the network regularization term has a fixed range, we also normalize $\mathbf{G}$ and $\mathbf{Y}$ so that the $Trace(\mathbf{Y} \cdot \mathbf{L}_G \cdot \mathbf{Y}^T)$ is within the range of $[0, 1]$ in the implementation of our model. More specifically, we set $\lVert \mathbf{G} \rVert_{F} = 1$, $\lVert \mathbf{Y}_{\cdot, i} \rVert = \frac{1}{\sqrt{p}}, i=1,2,\cdots, p$ (this also ensures that $\lVert \mathbf{Y} \rVert_{F} = 1$). This facilitates multi-view integration as all the network regularizers from multiple views are on the same scale.

\subsection{Multi-view Factorization AutoEncoder with network constraints}
Eq.~\ref{eq:mf_objective2} shows the objective for a single view. We can easily extend it to multiple views:

\begin{equation} \label{eq:mv_objective}
\begin{split}
\argmin_{\{\mathbf{\Theta_{e}}^{(v)}, \mathbf{Y}^{(v)}\}} & \sum_{v=1}^{V} \big( \left\Vert \mathbf{M}^{(v)} - Encoder(\mathbf{M}^{(v)}, \mathbf{\Theta_{e}}^{(v)}) \cdot \mathbf{Y}^{(v)} \right\Vert^2_F + \alpha \ Trace(\mathbf{Y}^{(v)} \cdot \mathbf{L}_{G^{(v)}} \cdot \mathbf{Y}^{(v)^T}) \big)
\end{split}
\end{equation}

Here for each of the $V$ views, we use a separate autoencoder. We combine all the reconstruction losses and feature interaction network regularizers together as the overall loss in Eq.~\ref{eq:mv_objective}. 

As mentioned before, $Encoder(\mathbf{M}^{(v)}, \mathbf{\Theta_{e}}^{(v)}) = \mathbf{X}^{(v)}$ can be regarded as learned latent factor representation for $N$ samples. Based on $\mathbf{X}^{(v)}$, we can derive patient similarity network $\mathbf{S}^{(v)}$ (which can also be used for spectral clustering). There are multiple ways to calculate a similarity network. Here we use cosine similarity as an example:

\begin{equation}\label{eq:sim_graph_cosine}
	S_{ij}=\frac{\lvert \mathbf{X}_{i, \cdot} \cdot \mathbf{X}_{j, \cdot} \rvert}{\lVert \mathbf{X}_{i, \cdot}\rVert \cdot \lVert \mathbf{X}_{j, \cdot} \rVert}
\end{equation}

For each view $v$, we get a patient similarity network $\mathbf{S}^{(v)}$ (Eq.~\ref{eq:sim_graph_cosine} omits the superscript for clarity). In addition, the outputs of multiple encoders can be combined.

\begin{equation}\label{eq:sum_x}
	\mathbf{X} = \sum_{v=1}^{V} \mathbf{X}^{(v)} = \sum_{v=1}^{V}Encoder(\mathbf{M}^{(v)}, \mathbf{\Theta_{e}}^{(v)})
\end{equation}

This idea is very much like ResNet \cite{he2016deep}. Another possible approach is simply concatenating all views together like DenseNet \cite{Huang2017}. We have tried using both in our experiments and the results are similar. We can then use the fused view $\mathbf{X}$ to calculate a patient similarity network $\mathbf{S}_X$ using Eq.~\ref{eq:sim_graph_cosine} again.

Since $\mathbf{S}_X, \text{ and } \mathbf{S}^{(v)}, v=1,2,\cdots, V$ are about the same set of patients and thus related to each other, we can fuse them together (this is a special case of affinity network fusion \cite{ma2017integrate}):

\begin{equation} \label{eq:complementary_sim_graph}
	\mathbf{S} = \frac{1}{V+1} \big(\sum_{i=1}^{V} \mathbf{S}^{(v)} + \mathbf{S}_X \big)
\end{equation}

Just like the feature interaction network regularizer (Eq.~\ref{eq:trace_y_L}), we can add a regularization term on view similarity:

\begin{equation} \label{eq:trace_X_S}
	Trace(\mathbf{X}^{(v)^T} \cdot \mathbf{L}_{S} \cdot \mathbf{X}^{(v)})
\end{equation}

$\mathbf{L}_{S}$ is the graph Laplacian of $\mathbf{S}$. 
Adding this term to Eq.~\ref{eq:mv_objective}, we get the new objective function: 
\begin{equation} \label{eq:mv_objective2}
\begin{split}
\argmin_{\{\mathbf{\Theta_{e}}^{(v)}, \mathbf{Y}^{(v)}\}} \sum_{v=1}^{V} \big(& \left\Vert \mathbf{M}^{(v)} - Encoder(\mathbf{M}^{(v)}, \mathbf{\Theta_{e}}^{(v)}) \cdot \mathbf{Y}^{(v)} \right\Vert^2_F \\ 
& + \alpha \ Trace(\mathbf{Y}^{(v)} \cdot \mathbf{L}_{G^{(v)}} \cdot \mathbf{Y}^{(v)^T}) 
 \\
& + \beta\	Trace(\mathbf{X}^{(v)^T} \cdot \mathbf{L}_{S} \cdot \mathbf{X}^{(v)}) \big)
\end{split}
\end{equation}

There are two kinds of networks involved in our framework: molecular interaction networks and patient similarity networks. For each type of -omic data, there is one corresponding interaction network $\mathbf{G}^{(v)}$. Unlike patient similarity networks, different molecular interaction networks involve different feature sets and cannot be directly merged. However, for patient similarity networks from multiple views, they are all about the same set of patients, and thus can fused to get a combined patient similarity network $\mathbf{S}$ using techniques such as affinity network fusion \cite{ma2017integrate}.

\subsection{Supervised learning with multi-view factorization autoencoder}

The proposed framework with the objective function Eq.~\ref{eq:mf_objective2} can be used for unsupervised learning (up to now, we have not used labeled data yet) with multiple view data and feature interaction networks available. When class labels or other target variables are available, we can apply the proposed model for supervised learning  by adding another loss term to Eq.~\ref{eq:mf_objective2}:

\begin{equation} \label{eq:mv_objective2_classification4}
\begin{split}
\argmin_{\{\mathbf{\Theta_{e}}^{(v)}, \mathbf{Y}^{(v)}\}} & 
\mathcal{L}(\mathbf{T}, \big(\sum_{v=1}^{V}Encoder(\mathbf{M}^{(v)}, \mathbf{\Theta_{e}}^{(v)})\big) \cdot \mathbf{C})\\ 
&+ \eta \sum_{v=1}^{V} \big( \left\Vert \mathbf{M}^{(v)} - Encoder(\mathbf{M}^{(v)}, \mathbf{\Theta_{e}}^{(v)}) \cdot \mathbf{Y}^{(v)} \right\Vert^2_F)\\
& + \alpha \sum_{v=1}^{V} Trace(\mathbf{Y}^{(v)} \cdot \mathbf{L}_{G^{(v)}} \cdot \mathbf{Y}^{(v)^T}) \\
& + \beta\ \sum_{v=1}^{V} Trace(\mathbf{X}^{(v)^T} \cdot \mathbf{L}_S \cdot \mathbf{X}^{(v)})
\end{split}
\end{equation}

The first part $\mathcal{L}(\mathbf{T}, \big(\sum_{v=1}^{V}Encoder(\mathbf{M}^{(v)}, \mathbf{\Theta_{e}}^{(v)})\big) \cdot \mathbf{C})$ is for either classification loss (e.g., cross entropy loss) or regression loss (e.g., mean squared error for continuous target variables). 
$\mathbf{T}$ is the true class labels or other continuous target variables available for training the model.

As in Eq.~\ref{eq:sum_x}, $\sum_{v=1}^{V}Encoder(\mathbf{M}^{(v)}, \mathbf{\Theta_{e}}^{(v)})$ refers to the sum of the last hidden layers of $V$ autoencoders (the output of the last hidden layer is also the encoder output). This represents the learned patient representations combining multiple views. $\mathbf{C}$ is the weights for the last fully connected layer typically used in neural network models for classification tasks.

The second part $\sum_{v=1}^{V} \big( \left\Vert \mathbf{M}^{(v)} - Encoder(\mathbf{M}^{(v)}, \mathbf{\Theta_{e}}^{(v)}) \cdot \mathbf{Y}^{(v)} \right\Vert^2_F)$ is the reconstruction loss for all the submodule autoencoders. The third and four term are the graph Laplacian constraints for molecular interaction networks and learned patient similarity networks as in Eq.~\ref{eq:trace_y_L} and Eq.~\ref{eq:trace_X_S}. $\eta, \alpha, \beta$ are non-negative hyperparameters adjusting the weights of the reconstruction loss, feature interaction network loss, and patient similarity network loss.

\begin{figure}
	\centering
	\includegraphics[width=0.8\linewidth]{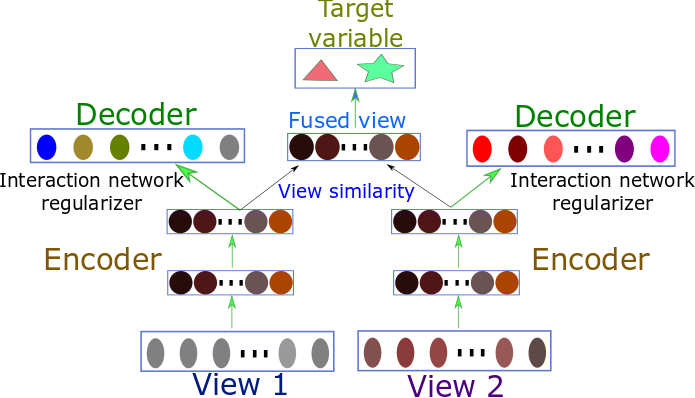}
	\caption{A simple illustration of proposed framework with two data views}
	\label{fig:model-architecture}
\end{figure}

The whole framework is end-to-end differentiable. 
A a simple illustration of the whole framework combining two views with two-hidden-layer autoencoders is depicted in Fig.~\ref{fig:model-architecture}.
We implement the model using PyTorch (\url{https://pytorch.org/}). Code will be made publicly available.

\section{Experiments}

\subsection{Dataset}
We downloaded the TCGA Pan-cancer dataset \cite{hutter2018cancer} and selected patients based on these criteria: 1) the patients' gene expression, miRNA expression, protein expression, and DNA methylation as well as clinical data are all available; and 2) the patients having the cancer types that have at least 100 patients. In total 6179 patients with 21 different cancer types were selected for analysis. 

\subsubsection{Target clinical variable}
We are trying to use four types of -omic data (i.e., gene expression, miRNA expression, protein expression and DNA methylation) to predict Progression-Free Interval (PFI) event and Overall Survival (OS) event. 
PFI and OS are derived clinical (binary) outcome endpoints \cite{liu2018integrated}. Both endpoints are relatively accurate, and are recommended to use for predictive tasks when available \cite{liu2018integrated}. PFI is preferred over OS given the relatively short follow-up time.

PFI=1 means the patient had a new tumor event in a fixed period, such as a progression of disease, local recurrence, distant metastasis,
new primary tumors, or died with the cancer without new tumor event. PFI=1 implies the treatment outcome is unfavorable.
PFI=0 means for patients without having a new tumor event in a fixed period or censored otherwise. 
There are 4268 patients with PFI=0 and 1911 patients with PFI=1.
OS=1 means for patients who were dead from any cause based on followup data; OS=0 for otherwise. There are 4460 patients with OS=0 and 1719 patients with OS=1. PFI and OS are the same for most cases (4941 out of 6179, or 80\%). As PFI is preferable to OS, we mainly use PFI as a binary target.

Since this is a highly unbalanced dataset and all the models including the baseline methods use prediction scores to decide binary labels, we report AUC (Area Under the ROC Curve) score as the main metric to evaluate classification performances. Other measures are similar to AUC but are less comprehensive.

\subsubsection{Data preprocessing}
For gene features, we performed log transformation and removed outliers. After filtering out genes with either low mean or low variance, 4942 gene features were kept for downstream analysis. For DNA methylation data, we removed features with low mean and variance. 4753 methylation features (i.e., beta values associated with CpG islands) were selected for analysis. For miRNA features, we also performed log transformation and removed outliers. For protein expression (RPPA) data, we removed nine features with NA values. 662 miRNA features and 189 protein features were kept for analysis. In total, there are 10,546 features from four -omic types.
For each of the four types of features, we normalize it to have zero mean and standard deviation equal to 1.

\paragraph{Molecular interaction networks}
We downloaded PPI database from STRING (v10.5) \cite{szklarczyk2014string} (\url{https://string-db.org/}). There are more than ten million protein-protein interactions with confidence scores between 0 and 1000. Since most interaction edges have a low confidence score, we selected about 1.5 million interaction edges with confidence scores at least 400. 
For gene and protein expression features, we extracted a subnetwork from this PPI interaction network. Since gene-gene interaction network is too sparse, we performed a one-step random walk (i.e., multiplying the interaction network by itself), removed outliers and normalized it. For miRNA and methylation features, we first map to miRNA/methylation to gene (protein) features, and then calculate a miRNA-miRNA and a methylation-methylation interaction network. Take miRNA data as an example. Let $\mathbf{M}_{mirna-protein}$ be the adjacency matrix for the miRNA-protein mapping (this matrix is derived from miRDB (\url{http://www.mirdb.org}) miRNA target prediction scores), and $\mathbf{M}_{protein-protein}$ be the protein-protein interaction network, then the miRNA-miRNA interaction network is calculated as follows:
\begin{equation*}
	\mathbf{M}_{mirna-mirna} = \mathbf{M}_{mirna-protein} \mathbf{M}_{protein-protein} \mathbf{M}_{mirna-protein}^T
\end{equation*}

We normalized all four feature interaction matrices so that their Frobenius norms are all equal to 1.

All the processed sample feature matrices and feature interactions matrices will be provided upon reasonable requests.

We randomly split the dataset into training set: 4326 patients, or 70\%;
validation set: 618 patients, or 10\%;
and test set: 1235 patients, or 20\%. We trained different models on the training set, and evaluated them on the validation set. We chose the model with the best validation accuracy to make predictions on test set, and reported the AUC score on test set. We shuffled the data and repeated the process ten times, and reported the average AUC as the final metric for model performances.

\subsection{Results}
We chose six traditional methods as well as plain neural network model as baselines. The six traditional methods are SVM, Decision Tree, Naive Bayes, kNN, Random Forest, and AdaBoost. 
Traditional models such as SVM only accept one feature matrix as input. So we used the concatenated feature matrix which has 6179 rows and 10,546 columns (features) as model input.
For kNN, we chose $k=5$ for all experiments. We used linear kernel for SVM. We used 10 estimators in Random Forest and 50 estimators in AdaBoost. 

For the plain autoencoder model with a classification head, we used a three-layer neural network. 
The input layer has 10,546 units (features). Both the first and second hidden layers have 100 hidden units. The last layer also has 10,546 units (i.e., the reconstruction of the input). We added a classification head which is a linear layer with two hidden units corresponding to two classes. 
This plain autoencoder model uses concatenated feature matrix as input and thus is view agnostic.

To facilitate fair comparisons, all of our proposed Multi-view Factorization AutoEncoder models share the same model architecture(i.e., two hidden layers each with 100 hidden units for each of the four submodule autoencoders), but the training objectives are different. 
Since this dataset has four different data types, our model has four autoencoders as submodules, each of which encodes one type of data (one view). Fig.~\ref{fig:model-architecture} shows our model structure (note in our experiments we have four views instead of only two shown in the figure). We combine the outputs of the four autoencoders (i.e., the outputs of the last hidden layers) by adding them together (Eq.~\ref{eq:sum_x}) for classification tasks.

The training objective for the Multiview Factorization AutoEncoder without graph constraints includes only the first two terms in  Eq.~\ref{eq:mv_objective2_classification4}.
The objective for the Multiview Factorization AutoEncoder with feature interaction network constraints (feat\_int) includes the first three terms in Eq.~\ref{eq:mv_objective2_classification4}. The objective for the Multiview Factorization AutoEncoder with patient view similarity network constraints (view\_sim) includes the first two and the last terms in Eq.~\ref{eq:mv_objective2_classification4}. And the objective for the Multiview Factorization AutoEncoder with both feature interaction and view similarity network constraints includes all four terms in Eq.~\ref{eq:mv_objective2_classification4}.

As our proposed model with network constraints is end-to-end differentiable, we trained it with Adam \cite{kingma2014adam} with weight decay $10^{-4}$. The initial learning rate is $5\times 10^{-4}$ for the first 500 iterations and then decreased by a factor of 10 (i.e., $5\times 10^{-5}$) for another 500 iterations. Models with the best validation accuracies are used for prediction on the test set.

The average AUC scores (10 runs) for predicting PFI and OS using these models are shown in Table.~\ref{tbl:auc_pfi_os}. Our proposed models (in bold font) achieved better AUC scores for both predicting PFI and OS. Note that traditional methods such as SVM do not perform as well as deep learning models. This may be due to the superior representation power of deep learning. Though our proposed Multi-view Factorization AutoEncoder is only slightly better than the plain autoencoder model, adding feature interaction and view similarity network constraints further improved the classification performance. Note that both \textit{Multi-view Factorization AutoEncoder + view\_sim} and \textit{Multi-view Factorization AutoEncoder + feat\_int + view\_sim} achieves the best AUC for PFI prediction. It seems adding both feature interaction and patient view similarity network constraints only improves the model performance very slightly. 
We suspect one main reason for this is because the dataset itself contains a lot of noise (due to the nature of multi-omic data) and the feature interaction networks derived from public knowledgebases are incomplete and noisy, too. If a larger dataset consisting hundreds of thousands of patients is available, we expect our proposed model with more network constraints to be able to generalize better.

\begin{table}
	\caption{AUC scores for PFI and OS}
	\label{tbl:auc_pfi_os}
	\centering
	\begin{tabular}{ccc}
		\toprule
		Model Name     & AUC (OS)    & AUC (PFI) \\
		\midrule
		SVM & 0.699  & 0.625     \\
		Decision Tree     & 0.670 & 0.634      \\
		Naive Bayes     & 0.655 &  0.644 \\
		kNN & 0.706 & 0.659 \\
		Random Forest & 0.720 & 0.661\\
		AdaBoost & 0.716 & 0.689\\
		Plain AutoEncoder Model & 0.758 & 0.716 \\
		\textbf{Multi-view Factorization AutoEncoder without graph constraints} & 0.761 & 0.717\\
		\textbf{Multi-view Factorization AutoEncoder + feat\_int} & 0.765 & 0.721\\
		\textbf{Multi-view Factorization AutoEncoder + view\_sim} & 0.763 & \textbf{0.724}\\
		\textbf{Multi-view Factorization AutoEncoder + feat\_int + view\_sim} &\textbf{0.766} & \textbf{0.724}\\
		\bottomrule
	\end{tabular}
\end{table}

In addition, we had tried to use DenseNet \cite{Huang2017} and ResNet \cite{he2016deep} as the backbone of the autoencoders instead of multi-layer perceptrons, and we experimented with different number of hidden units and hidden layers. Using DenseNet as the backbone with three hidden layers each with 100 units achieved best AUC scores (0.725) for predicting PFI. But the results are not significantly different and thus not presented here.

\subsubsection{Learned feature embeddings preserve interaction network structure}
\begin{figure}
	\centering
	\includegraphics[width=0.8\linewidth]{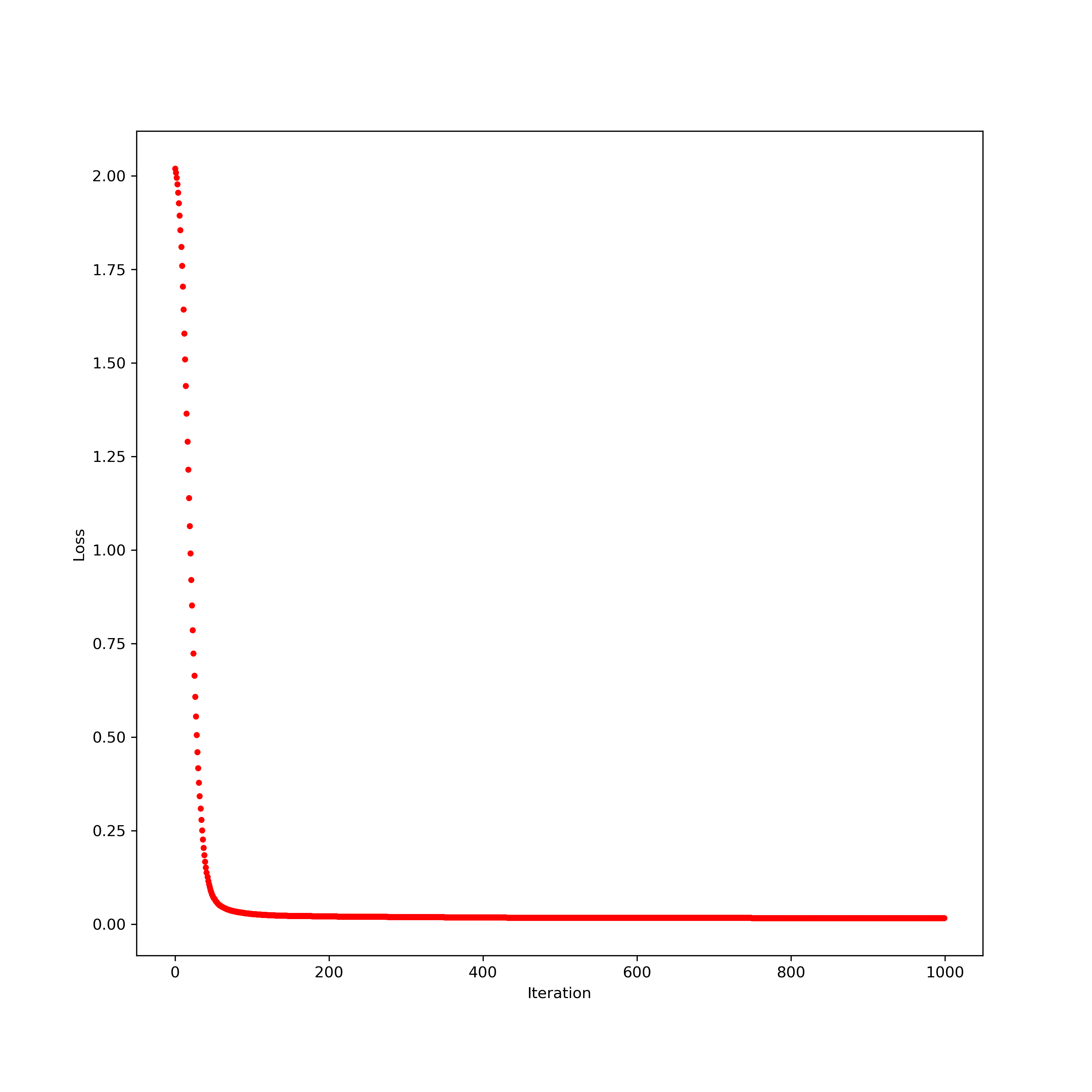}
	\caption{A typical feature interaction network regularizer training loss curve}
	\label{fig:loss-feat-int}
\end{figure}

Our proposed model learns patient representations and feature embeddings simultaneously. While patients are different from datasets to datasets, the genomic features (such as gene features) and their interaction networks are from domain knowledge, and thus are persistent regardless of which dataset we are using. Since we have a regularization term in the loss to ensure the learned feature embeddings are consistent with feature interaction networks, we would like to know if the model is able to learn an embedding that is ``compatible'' with the domain knowledge of interaction networks. 
We plotted the loss term $\sum_{v=1}^{V} Trace(\mathbf{Y}^{(v)} \cdot \mathbf{L}_{G^{(v)}} \cdot \mathbf{Y}^{(v)^T})$ from one typical run of training our model with feature interaction network constraints in Fig.~\ref{fig:loss-feat-int}. This regularization term decreased to nearly zero very fast, which means the information from feature interaction networks is fully assimilated into the model, or more specifically, the weights of the decoders in the model. We found that many independent runs show very similar loss curves, which means the model is able to robustly learn an feature embedding that preserves the feature interaction network information.

\section{Conclusion}
Multi-omic integrative analysis is important for cancer genomics. While multi-omic data has the ``big p, small N'' problem, biological knowledge can be used as a leverage for large-scale data integration and knowledge discovery. A number of databases such as STRING \cite{szklarczyk2014string}, Reactome Pathways \cite{croft2013reactome}, etc., can be used to extract biological interaction networks. Intelligently integrating these biological networks into a model is crucial for mining multi-omic data. We proposed the Multi-view Factorization AutoEncoder Model with network constraints to integrate multi-omic data and molecular interaction networks for multi-omic data analysis. 
 
Our model contains multiple factorization autoencoders as submodules for different views, and combines multiple views with their high-level latent representations. The factorization autoencoder utilizes a deep architecture for the encoder and a shallow architecture for the decoder. This on one hand increases the overall model representation power, on the other provides a natural way to integrate graph constraints into the model. 

Our model learns patient embeddings and feature embeddings simultaneously, enabling us to add network constraints on both feature interaction networks and patient view similarity networks. 
Our approach can be applied to large-scale multi-omic dataset to learn embeddings for molecular entities, subject to network constraints that ensure the learned representations are consistent with molecular interaction networks. Meanwhile the model can produce patient representations for each view. As the latent patient representations from multiple views should be similar to each other, we added a network regularizer to encourage the learned patient representations in multiple views to be consistent with one another with respect to patient similarity networks. 

The experimental results on the TCGA pan-cancer dataset show that our proposed model with feature interaction network and patient view similarity network constraints outperforms other traditional methods and plain deep learning autoencoder models. Though we mainly focused our discussion on multi-omic data analysis, as a general approach, our proposed method can be applied to any other multi-view data with feature interaction networks. Future work may focus on using more advanced autoencoders (e.g., adversarial autoencoder \cite{makhzani2015adversarial}) and encoding some domain knowledge directly into the model architecture.



\end{document}